\documentclass[10pt,twocolumn,letterpaper]{article}

\usepackage{cvpr}              % To produce the CAMERA-READY version
\usepackage{graphicx}
\usepackage{amsmath}
\usepackage{amssymb}
\usepackage{booktabs}
\usepackage{multirow}

\usepackage[pagebackref,breaklinks,colorlinks]{hyperref}

\usepackage[capitalize]{cleveref}
\crefname{section}{Sec.}{Secs.}
\Crefname{section}{Section}{Sections}
\Crefname{table}{Table}{Tables}
\crefname{table}{Tab.}{Tabs.}

\def\cvprPaperID{xxxx} % *** Enter the CVPR Paper ID here
\def\confName{CVPR}
\def\confYear{2022}

\begin{document}
%%%%%%%%% TITLE - PLEASE UPDATE
\title{Exploring and Evaluating Image Restoration Potential in Dynamic Scenes}
% A New Clue for Computer Vision: Image Restoration Potential
% A New Clue in Dynamic Scenes: Image Restoration Potential

% \author{First Author\\
% Institution1\\
% Institution1 address\\
% {\tt\small firstauthor@i1.org}
% % For a paper whose authors are all at the same institution,
% % omit the following lines up until the closing ``}''.
% % Additional authors and addresses can be added with ``\and'',
% % just like the second author.
% % To save space, use either the email address or home page, not both
% \and
% Second Author\\
% Institution2\\
% First line of institution2 address\\
% {\tt\small secondauthor@i2.org}
% }

\author{Cheng Zhang$^{\dag ,1}$, Shaolin Su$^{\dag ,1}$, Yu Zhu$^{\ddag ,1}$, Qingsen Yan$^2$, Jinqiu Sun$^1$, Yanning Zhang$^1$
\thanks{$\dag$~denotes equal contribution. 
This work is supported by National Science Foundation of China under Grant No. U19B2037, 61901384, Natural Science Basic Research Program of Shaanxi No. 2021JCW-03 and National Engineering Laboratory for Integrated Aero-Space-Ground-Ocean Big Data Application Technology. $^{\ddag}$ Corresponding author: Yu Zhu.}
\\$^1$School of Computer Science and Engineering, Northwestern Polytechnical University, China
\\$^2$School of Computer Science and Engineering, The University of Adelaide, Australia
\\
{\tt\small https://github.com/Justones/IRP}
}

\maketitle

\begin{abstract}
   In dynamic scenes, images often suffer from dynamic blur due to superposition of motions or low signal-noise ratio resulted from quick shutter speed when avoiding motions. Recovering sharp and clean results from the captured images heavily depends on the ability of restoration methods and the quality of the input. Although existing research on image restoration focuses on developing models for obtaining better restored results, fewer have studied to evaluate how and which input image leads to superior restored quality. In this paper, to better study an image's potential value that can be explored for restoration, we propose a novel concept, referring to image restoration potential (IRP). Specifically, We first establish a dynamic scene imaging dataset containing composite distortions and applied image restoration processes to validate the rationality of the existence to IRP. Based on this dataset, we investigate several properties of IRP and propose a novel deep model to accurately predict IRP values. By gradually distilling and selective fusing the degradation features, the proposed model shows its superiority in IRP prediction. Thanks to the proposed model, we are then able to validate how various image restoration related applications are benefited from IRP prediction. We show the potential usages of IRP as a filtering principle to select valuable frames, an auxiliary guidance to improve restoration models, and even an indicator to optimize camera settings for capturing better images under dynamic scenarios.

\end{abstract}

%%%%%%%%% BODY TEXT
\section{Introduction}
\label{sec:intro}

\begin{figure}[t]
\centering
\includegraphics[angle=0,width=0.5\textwidth]{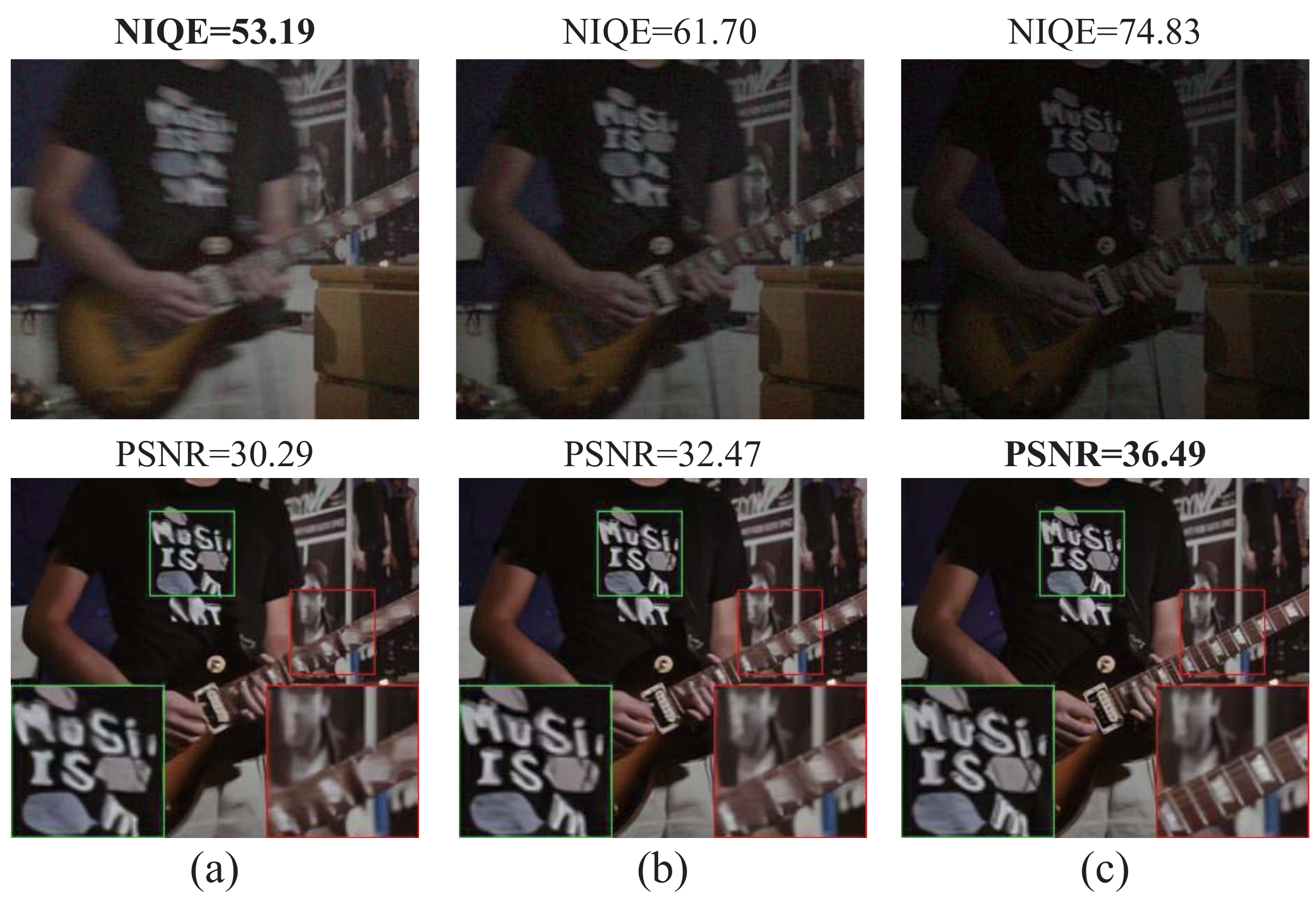}
\vspace{-0.6cm}
\caption{We show how the proposed concept IRP differs from the traditional image metric visual quality. Given a sequence of images captured under varying exposures in dynamic scenes (first row), both IQA metric NIQE \cite{mittal2012making} and human observers favorite either (a) or (b) as relatively good shots, but they do not necessarily lead to better restored results (second row). As a comparison, a perceptually poor image (c) leads to the best restored image quality. The result indicates the necessity of developing the IRP metric which predicts if the restored image quality will be good even before it is processed. All images are processed by the same restoration model MPR\cite{zamir2021multi} to ensure fairness.}
\label{fig:1}
\vspace{-0.6cm}
\end{figure}

In real world imaging scenarios with dynamic motions, degradation is a common factor due to moving objects or shaking devices. To avoid blur, the photographer can shorten the exposure time, but results in low illumination regions and notorious noises \cite{chen2018learning, wei2020physics, chang2021low}. With varying camera settings, the type of distortion may change, but hardly diminishes.
To alleviate the annoying distortions, different image restoration algorithms are required: 
one can apply either deblurring approaches to remove motion blur in the image captured from adequate exposure, or denoising methods to alleviate noise artifacts due to shortened exposure. However, among the noise-blur trade-off, different kinds of distorted input lead to different restored results.
Naturally, a question comes up that under the trade-off, which kind of image leads to better restored results? The question is fundamental, yet has not been well investigated in the literature. As far as we know, in order to obtain better restored results, most researches focus on developing restoration algorithms \cite{zamir2021multi,chen2021hinet,guo2019toward,gu2014weighted,zamir2020learning,gong2017motion}, but few are carried out to evaluate how and which input images lead to superior restored quality.
As restored image quality heavily depends on both restoration models and inputs, in this paper, we propose the concept of image restoration potential (IRP), denoting an inherent image attribute that measures the potential value of an image that can be explored for restoration.

Though intuitively, less distorted images with higher quality lead to better restored results, the quality metric here differs from the widely accepted concept of visual perceptual quality. As shown in the top row of Figure \ref{fig:1}, giving a sequence of images captured in dynamic scene, both image quality assessment (IQA) metric NIQE \cite{mittal2012making} and human observers tend to favorite image (a) or (b) as relatively better shots, however, these ``better'' inputs do not necessarily lead to better restored ones. In contrast, a perceptually poor quality image (c) achieves the best restored result, shown in the bottom row. The phenomenon therefore inspired us the necessity of developing the IRP concept. As traditional IQA metrics\cite{su2020blindly,Bovik2015ILNIQE} aim at measuring image quality at the \textbf{present} view and extracting perceptual features which are adapted to the human visual system (HVS), IRP is proposed to forecast if the restored image quality will be good \textbf{even before} they are processed, and focus more on the degradation that closely relates to the restoration process.

To investigate the proposed IRP, we first established a dynamic scene imaging dataset by simulating 5500 degraded images captured under various camera settings. We then conducted 4 representative restoration algorithms on each of the images in the dataset, and validated that IRP belonging to an inherent image attribute regardless of how concrete restoration algorithms are applied. We further dig into the distortions that exist in dynamic imaging scenarios including noises, blurriness, and low illumination, and proposed a deep model for accurate IRP prediction. By disentangling distortion factors and selective fusing degradation features, the proposed model showed its effectiveness in revealing image potentials for restoration. Lastly, we apply IRP prediction to various kinds of applications, including filtering valuable frames in image sequences for efficiently processing, guiding image processing models for adaptive restoration, and optimizing camera settings for capturing images leading to better restored quality.

To summary, the contributions of this paper include:

\begin{itemize}
\item
We introduce a novel image attribute, named IRP, as a criterion to measure the potential value of an image that can be explored for restoration. By collecting 5500 images as well as their restored quality labels, we investigate several properties of the proposed IRP.

\item
We analyze the key factors affecting IRP and develop a deep model for IRP prediction. By gradually distilling image distortions that exist in dynamic scenes and selective fusing the features to form complement representations, we verify the superior prediction accuracy of the proposed model.

\item
We show potential usages of the proposed IRP to various image restoration related applications. IRP has shown its effectiveness in filtering valuable frames for the restoration process, providing auxiliary guidance to restoration models, and even optimizing camera settings when capturing images under dynamic scenarios.

\end{itemize}

\section{Related work}
\subsection{Image Restoration}

According to the leading type of degradation contained in an image, different kinds of restoration tasks are proposed. In dynamic scenes, when the camera exposure setting is set long to ensure sufficient light, motions will be apparent and deblurring methods are required. Representative models include conventional approaches\cite{xu2013unnatural,whyte2012non} and CNN based models \cite{gong2017motion,nah2017deep,kupyn2018deblurgan,zhang2019deep,wang2021non,zamir2021multi,chen2021hinet}. Meanwhile, when exposure is set short to avoid the superposition of motions, noise artifacts will be obvious due to insufficient light, and denoising approaches are needed, including conventional techniques \cite{aharon2006k,dabov2007image,gu2014weighted,xu2017multi} and CNN based methods \cite{zhang2017beyond,guo2019toward,anwar2019real,zhang2020attention,zamir2021multi}.
The above approaches, though achieving improving performances in their own field, are however proposed to deal with separate image restoration tasks. As a comparison, in this paper, we aim at developing a general image measurement IRP, that adapts to both restoration problems, and we are interested in finding out how image restoration related tasks are benefited from IRP predictions.

\subsection{Image Quality Assessment}

The goal of IQA is to enable machines to perceive the visual quality of images, being consistent with human perceptual results. By assessing an image's visual quality, many vision related tasks could be quantitatively measured and potentially optimized. Existing IQA approaches including full-reference IQA \cite{Zhou2004Image,wang2010information,zhang2018unreasonable}, reduced-reference IQA \cite{xu2015fractal,liu2017reduced,golestaneh2016reduced}, and no-reference IQA \cite{Mittal2012BRISQUE, su2020blindly, zhu2020metaiqa}, according to the accessibility to the pristine reference image. 
Though many works have been carried out in learning the relationship between image features and their direct visual quality \cite{Bovik2015ILNIQE, Xu2016Blind, su2021koniq++}, none of the work has studied an image's potential quality that can be explored for restoration. Therefore, in this paper, we propose and investigate the concept of IRP, analyze its properties, show its differences to IQA, and evaluate its potential usages in real world applications.

\section{Investigating on IRP}

To investigate IRP, we first established a dataset called Dynamic Scene-IRP (DS-IRP). The collection of DS-IRP mainly includes two stages. In the first stage, we collected 2,500 dynamic scenes and 11 images in each scene under various camera settings. Since in dynamic scenes, it is impractical to acquire ground truth images by shooting in reality, which are however required in stage 2, we thus chose to simulate images following a dynamic imaging formation and finally had 27,500 images in total.
In the second stage, we re-trained and tested 4 representative restoration algorithms on each of the images collected from stage 1. The IRP labels are then acquired by calculating restored image quality by referencing ground truth images. With the established DS-IRP dataset, we investigated IRP and revealed several of its properties.

\subsection{Dynamic imaging formation}

In our dynamic imaging formation, we model the imaging process starting with scene radiance $\phi$ and ending with image value $I$ in sRGB space, but paying particular attention to dynamic motions existing in scenes. Therefore, the overall dynamic imaging formation can be expressed as the joint result including radiant power $\phi$, motion information $m$ and noise $n$ during the exposure time $\Delta t$:

\vspace{-0.2cm}
\begin{equation}\label{equation1}
    I = G(\phi,m,n,\Delta t)
\end{equation}
\vspace{-0.45cm}
% where $G$ and $I$ represent the single-shot imaging function model and image value in sRGB space.

Specifically, we consider motions during exposure, then convert the scene radiance into linear RGB pixel values, following \cite{chang2021low,dahary2021digital,onzon2021neural}:
\vspace{-0.2cm}
\begin{equation}\label{equation2}
  I_{l} =  y_{p}(\int_{t_0}^{t_0+\Delta t} \phi_t m_t\, dt) + n
\end{equation}
where $y_{p}$ converts the received photoelectrons expressed by exposed photosite during the exposure time $\Delta t$ into voltages, which are further recorded by sensor as linear RGB pixel values. $m_t$ denotes motion at $t$ time, and $n$ represents the overall noise.

In practice, motions are considered in an equivalent but simpler way after $\phi$ are converted into linear RGB signals. Following \cite{gong2017motion}, we apply optical flow to linear RGB signals as motion information. Specifically, we select consecutive image frames containing real scene motions from the vimeo-triplet dataset\cite{xue2019video} and estimate the optical flow by ARflow\cite{liu2020learning}, to represent motions $m_{t_0}$ that exist in dynamic scenes. To cover various dynamic imaging results under different camera exposure settings, we sample 11 diverse exposure times in each scene and scale the motions $m_{t_0}$ according to exposure time. In total, 2,500 scenes are selected from the vimeo-triplet dataset, each containing 11 imaging results corresponding to various exposure times. In each scene, we also collect the original image from the vimeo-triplet dataset, which serves as the ground truth image that is used for generating IRP labels in the second stage.

In the dynamic imaging formation, we consider noise as another inevitable degradation factor. The overall noise $n$ composes the combination of shot noise $n_{shot}$ and readout noise $n_{read}$, formulated as:

\vspace{-0.3cm}
\begin{equation}
    n = n_{shot} + n_{read}
\end{equation}
where $n_{shot}$ originates from the particle nature of light, and follows a Poisson process as $(I_l + n_{shot}) \sim Poisson(I_l)$ \cite{wei2020physics}, while $n_{read}$ associates with voltage fluctuation in signal processing flow, and follows a zero-mean Gaussian distribution with device-specific standard deviation, termed as $n_{read} \sim Gaussian(0, \sigma ^{2}) $.

Finally, image signals are converted from linear RGB to sRGB space by quantization and bounded transformation, expressed as:
\vspace{-0.2cm}
\begin{equation}
  I = min(\lfloor (I_{l})^{\gamma} + 0.5 \rfloor,M_{max})
\end{equation} 
where $\lfloor \cdot \rfloor$, $\gamma$ and $M_{max}$ indicate floor function, gamma transformation, and the maximum value recorded by the camera sensor respectively.

\subsection{IRP Label Generation}
\label{sec:3.2}

After obtaining 27,500 dynamic imaging results, we further collect their IRP labels. Since IRP is proposed to measure the potential that an image can be explored for restoration, we thus first apply image restoration algorithms on the images, then represent IRP value as the restored image quality, which is calculated by referencing the ground truth. 

During IRP generation procedure, we select four representative image restoration methods, being effective in processing diverse image distortions, including Unet\cite{ronneberger2015u}, MIRnet\cite{zamir2020learning} ,MPR\cite{zamir2021multi} and HInet\cite{chen2021hinet} for restoration. Since we collected images under 11 diverse exposure settings, we train and test each restoration model under separate exposure settings across all scenes, leading to a total $4\times 11 = 44$ times of training and testing procedures. Though it is plausible to mix all images for training and testing, we found that the models become less capable of handling all mixing distortions from noise to blur integrated with one model alone. As a result, we chose to train and test restoration models under each individual exposure settings, expecting that the max amount of ``restoration potential'' could be explored by the models. Specifically, among 2,500 scenes, we take 2,000 scenes for training restoration models, leaving the rest 500 scenes for testing and obtaining restoration results. To generate IRP values, we calculate both PSNR and LPIPS \cite{zhang2018unreasonable} scores between restored images and the ground truth to balance the trade-off between distortion and perception \cite{blau2018perception}. After normalizing both criteria to the range of $[0,1]$, we calculate their mean value as IRP measurement. At last, for each image, we average the 4 IRP values corresponding to 4 restoration methods as the final IRP score. The whole generation process collects $500 \times 11= 5500$ IRP labels in total. The labels, along with corresponding images collected in stage 1, form the DS-IRP dataset. Examples of data and IRP labels can be found in our supplementary.

\begin{figure}[thbp]
\centering   %图片居中排列
\subfloat{%前面中括号里面的a是子图标题
\begin{minipage}[t]{0.23\textwidth}%每个图形大小
   \centering
  \includegraphics[angle=0,width=1\textwidth]{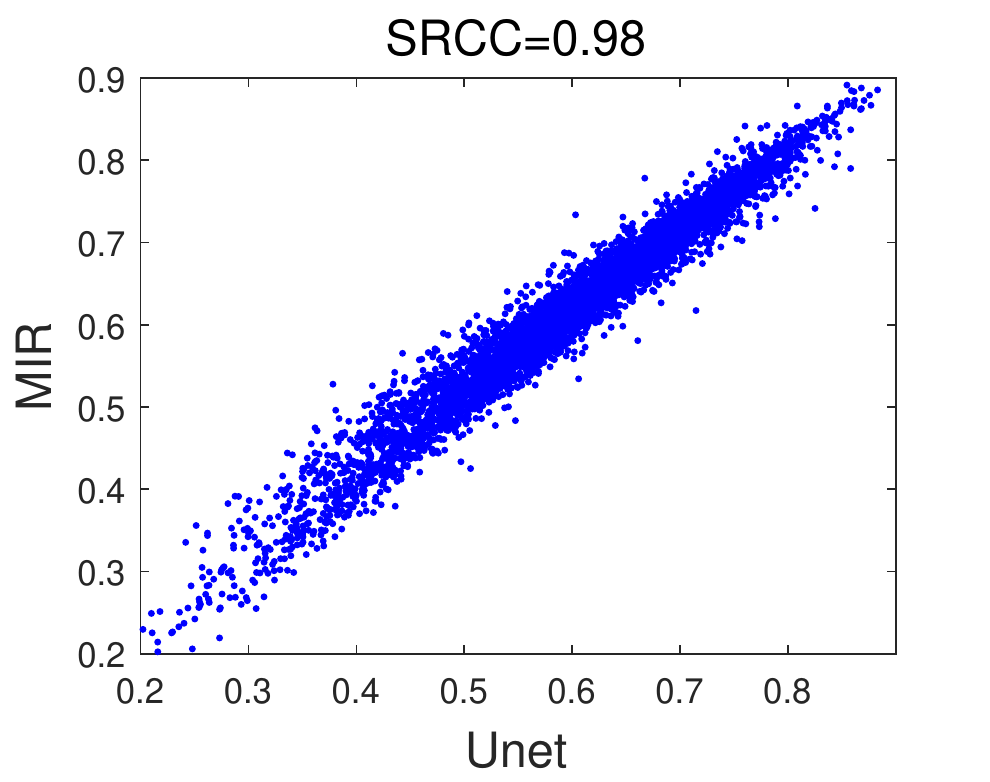}%插入图片,括号里面是图片路径
\end{minipage}
}
\hspace{-0.2cm}
\subfloat{
\label{fig:subfig_b}
\begin{minipage}[t]{0.23\textwidth}
   \centering
  \includegraphics[angle=0,width=1\textwidth]{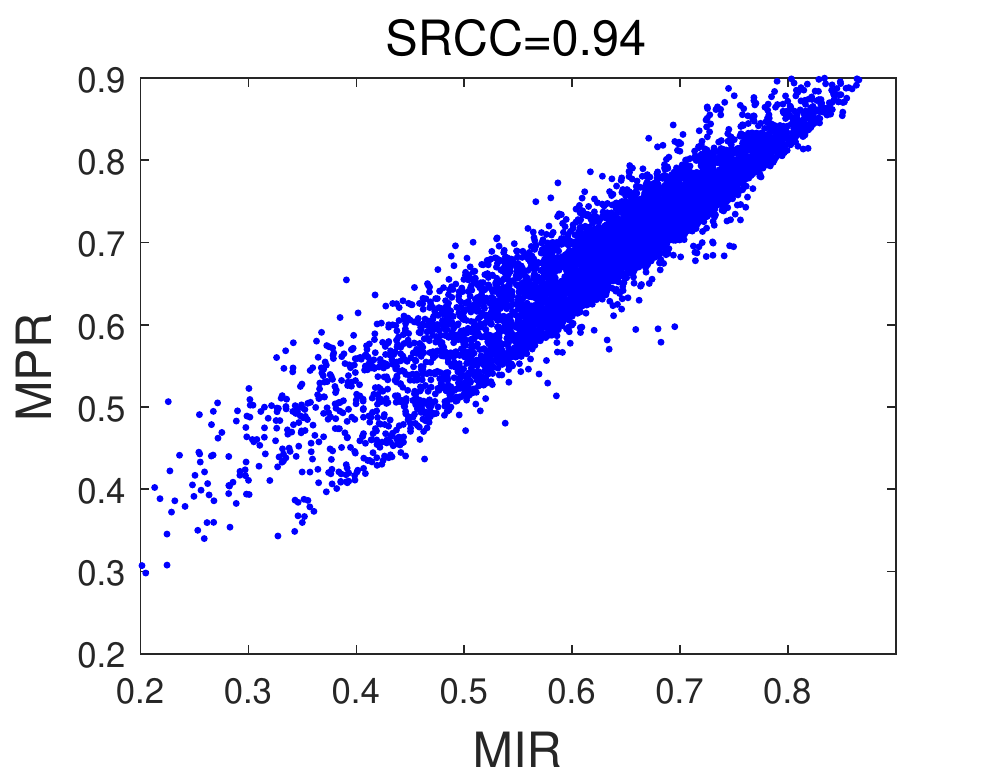}
\end{minipage}
}
\\%注意此处的换行不能省。
\hspace{-0.2cm}
\subfloat{
\label{fig:subfig_d}
\begin{minipage}[t]{0.23\textwidth}
   \centering
  \includegraphics[angle=0,width=1\textwidth]{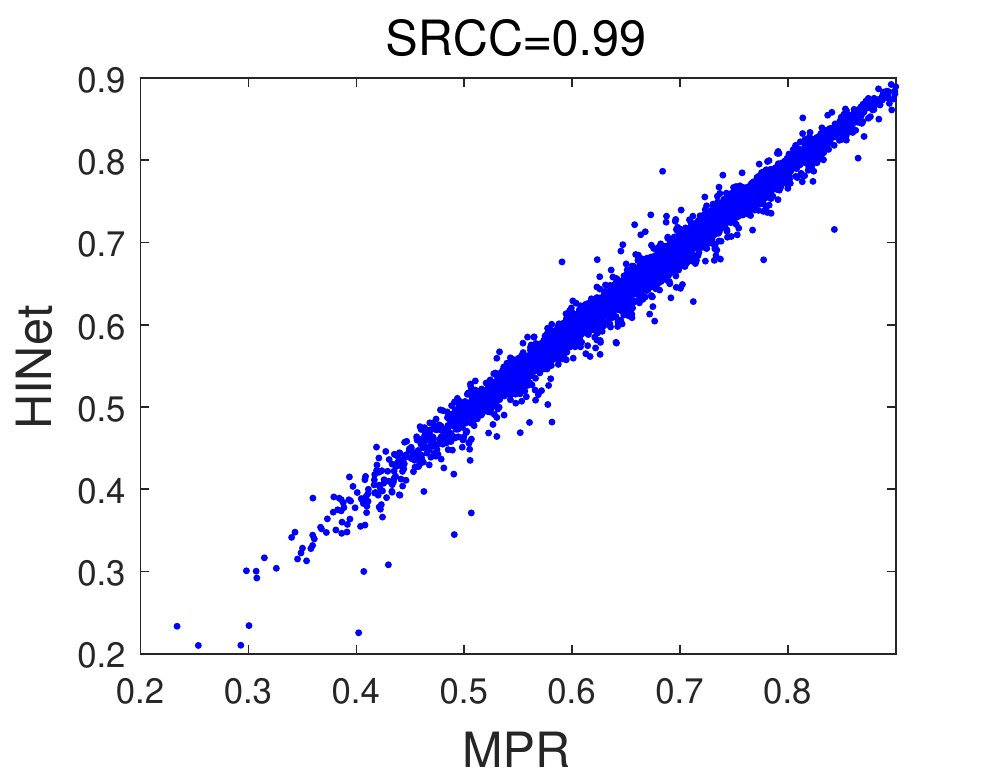}
\end{minipage}
}
\hspace{-0.3cm}
\subfloat{
\label{fig:subfig_e}
\begin{minipage}[t]{0.23\textwidth}
   \centering
  \includegraphics[angle=0,width=1\textwidth]{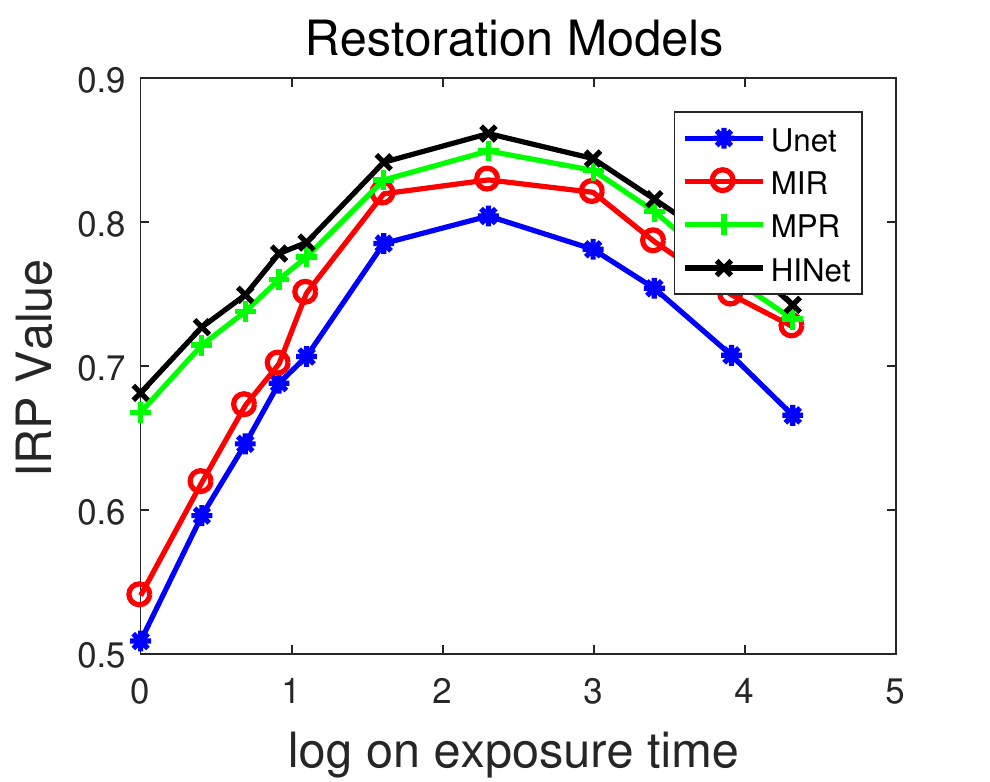}
\end{minipage}
}
\caption{In the first three subfigures, we plot IRP values between two arbitrary restoration models. The restored image quality is highly correlated across different models, indicating IRP as an inherent image attribute. In the fourth subfigure, we plot IRP values generated by four restoration models under one scene, the IRP values corresponding to four methods also keep consistent.}%图的大标题
\vspace{-0.6cm}
\label{fig:3.3.1}%所有图的引用标号，注意必须放在\caption后面才可以利用。
\end{figure}

\subsection{Investigation on IRP Properties}
After establishing the DS-IRP dataset, we are now able to investigate several properties of IRP. Specifically, we explored how IRP is affected by the selection of restoration models, how distortion factors determine IRP values and the difference between IRP and image visual quality.

\noindent
\textbf{Influence by Restoration Models.} Apparently, even for the same image, the selection of different restoration models results in different restored image quality. Therefore, we are interested in finding out how IRP is influenced by the selection of restoration models. 
To this end, we take IRP values generated under different restoration models to compare their correlations, and we show the results in Figure \ref{fig:3.3.1}. From the first three subfigures, it can be observed that despite different restoration models being applied, the restored image quality is highly correlated across all models. In the fourth subfigure, we show IRP values generated by four restoration models in one scene, containing 11 images captured under varying exposure settings. As directly shown, for each image, the relative IRP values keep consistent across restoration models.
The above results indicate that even if different restoration models are used, a good image is a good image, \emph{i.e.} its potential that can be explored for restoration keeps consistent. The result also demonstrates that IRP belongs to an inherent image attribute which is majorly mapped from image appearance, similar to existing image attributes including visual quality, brightness, and sharpness, \emph{etc}.

\begin{figure}[t]
\centering   %图片居中排列
\subfloat{%前面中括号里面的a是子图标题
\begin{minipage}[t]{0.165\textwidth}%每个图形大小
   \centering
  \includegraphics[angle=0,width=1\textwidth]{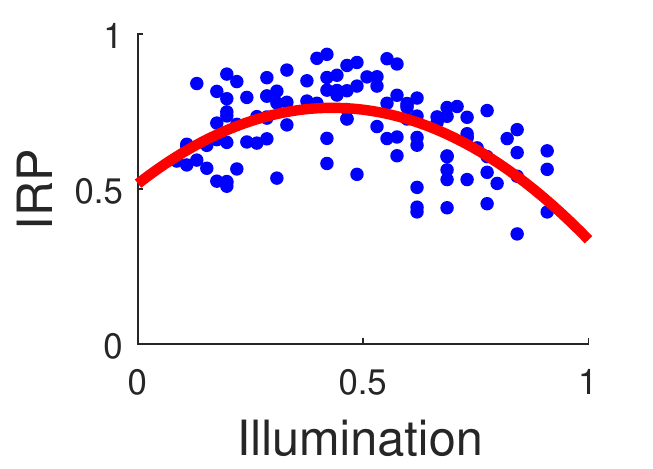}%插入图片,括号里面是图片路径
\end{minipage}
}
\hspace{-0.5cm}
\subfloat{
\begin{minipage}[t]{0.165\textwidth}
   \centering
  \includegraphics[angle=0,width=1\textwidth]{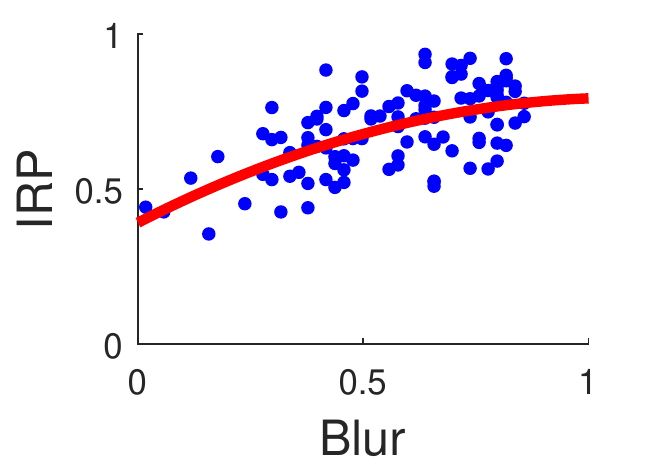}
\end{minipage}
}
\hspace{-0.5cm}
\subfloat{
\begin{minipage}[t]{0.165\textwidth}
   \centering
  \includegraphics[angle=0,width=1\textwidth]{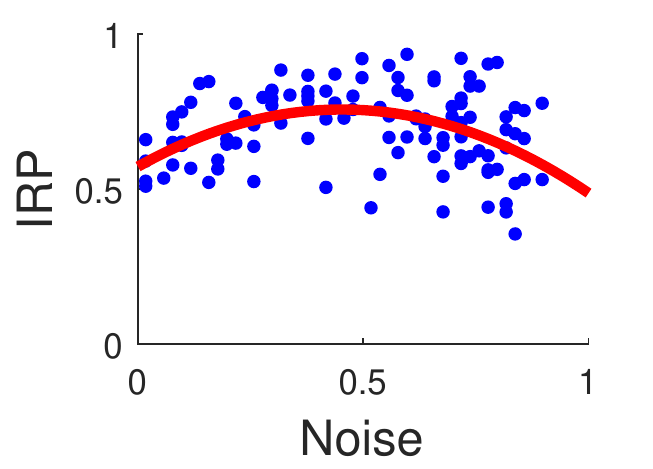}
\end{minipage}
}
\caption{Subjective study on relationship between degradation factors and IRP values.}%图的大标题
\label{fig:3.3.2}
\end{figure}

\noindent
\textbf{Analyse on Determinant IRP Factors.} As shown in Equation (\ref{equation1}), the dynamic imaging process is determined by scene radiance $\phi$, motion $m$, noise $n$ and exposure time $\Delta t$, reflected as illumination, blurriness and noise on image appearances. Therefore, we study on the three factors to analyze how they affect IRP values. Concretely, we conducted a subjective user study on three factors, by labeling the degradation magnitudes corresponding to low illumination, blur, and noise in 110 images randomly selected from the DS-IRP dataset. We then plot the relationship between each degradation magnitude and IRP value in Figure \ref{fig:3.3.2}. In our experiment, 40 observers are invited to rate the degradation level ranging from $[0, 1]$, where a larger rating indicates less degradation detected. From Figure \ref{fig:3.3.2}, we observe that when illumination and noise problem is severe, IRP values correlate with the magnitude of distortion. However, when two degradations are becoming less, IRP does not increase accordingly, due to the increasing motion blur resulted from longer exposure time in most scenes. This indicates when camera exposure setting varies, different kinds of distortions are becoming determinant under composite degradation. Meanwhile, we also found that when image blur is diminishing, IRP value consistently improves, suggesting that blurriness always plays a determinant role for IRP.

\begin{figure}[t]
\vspace{-0.5cm}
\centering   %图片居中排列
\subfloat{%前面中括号里面的a是子图标题
\begin{minipage}[t]{0.165\textwidth}%每个图形大小
   \centering
  \includegraphics[angle=0,width=1\textwidth]{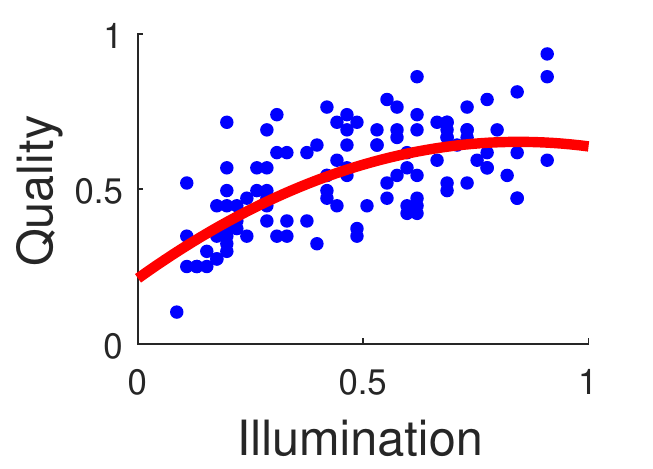}%插入图片,括号里面是图片路径
\end{minipage}
}
\hspace{-0.5cm}
\subfloat{
\begin{minipage}[t]{0.165\textwidth}
   \centering
  \includegraphics[angle=0,width=1\textwidth]{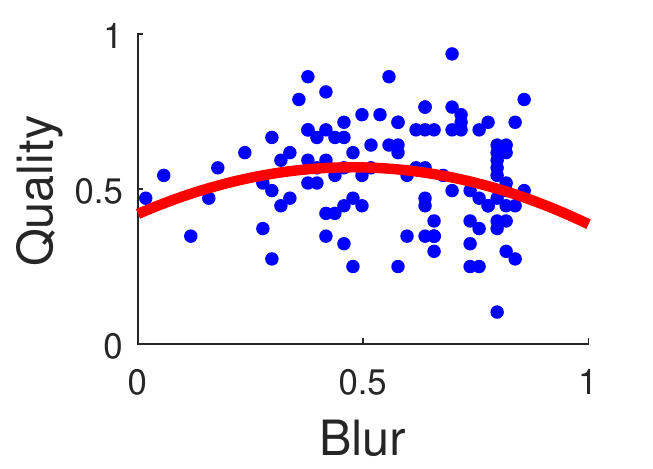}
\end{minipage}
}
\hspace{-0.5cm}
\subfloat{
\begin{minipage}[t]{0.165\textwidth}
   \centering
  \includegraphics[angle=0,width=1\textwidth]{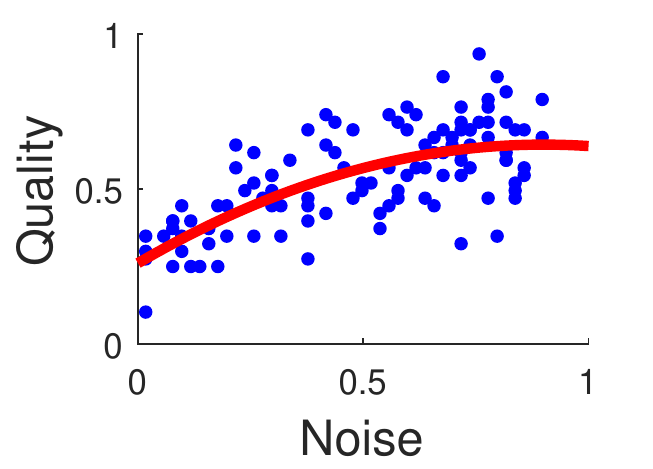}
\end{minipage}
}
\caption{Subjective study on relationship between degradation factors and image visual quality.}%图的大标题
\vspace{-0.6cm}
\label{fig:3.3.3}
\end{figure}

\noindent
\textbf{Comparison between IRP and Image Visual Quality.} We further compare IRP with the widely studied image attribute, \emph{i.e.} image visual quality, to investigate their differences. We conducted a similar user study by labeling image quality annotations, and plotting the relationship between each degradation factors and visual quality labels in Figure \ref{fig:3.3.3}. Compared with Figure \ref{fig:3.3.2}, it can be easily observed that image degradations, including low illumination, blurriness and noise, affect image visual quality in a different way from IRP. As can be seen, when composite distortions vary, visual quality shows a monotonic correlation with illumination and noise problems, while less correlated to blurriness. According to the above observation, we conclude the differences between IRP and visual quality: under the blur-noise trade-off in dynamic scenes, illumination problem attributes to a major factor determining image visual quality, but a relatively ``easy'' distortion for IRP as the information could be easily restored. Noise problem also implies relatively minor affects on IRP attribute, but is determinant to visual quality to some degree. Among the degradation factors, blurriness correlates most with IRP value, and serves as the ``hardest'' distortion for restoration.

\section{Learning to Predict IRP}

In this section, we aim at developing a deep model for accurately predicting IRP. The predicting model is expected to serve as an indicator being applied in various dynamic scene imaging or restoration tasks. As we have analysed the major three factors and their mutual interactions in affecting IRP values, we thus propose to gradually distill each of the three factors in our model by individual branches, in order to learn complimentary IRP representations. In addition, to better understand the impact from each kind of degradation among the composite distortions, we propose to selective fuse the degradation features, which are finally regressed to IRP scores by multi-layer perceptrons (MLP).

\subsection{Gradually Distilling Degradation Components}

Given input images contaminated by the composition of illumination, noise and dynamic blur problems, we propose to gradually distill the degradation components by a series of pre-processing techniques, and extract each component's features one by one.

As image illumination mainly affects image visual presentation but a ``easy to eliminate'' component for IRP, we thus use the original distorted image to extract illumination features. Specifically, we adopt image histograms in an individual branch to extract illumination statistics. We compute 256 bins of the histogram and apply 1 layer of 1D convolution with a kernel size of 7 to the bins. The histogram is then spatially expanded to fit the size of features extracted from the rest two branches, denoted as $\mathbf{F_i}$. Next, to distill noise and blur features from composite distortions, we scale image signal in linear RGB space to alleviate the effect of illuminations, and extract noise features by another branch. Inside the noise feature extraction branch, we use the feature extractor from stage 1 of the ResNet50 backbone \cite{he2016deep} to extract low level features. The features are then fed into 3 ASPP blocks \cite{chen2017deeplab} to expand the receptive field. In this way, we extract image features through a shallow branch, which leans to learn low level noise features, denoted as $\mathbf{F_n}$. As last, to distill blur features, we  apply guided filtering \cite{he2010guided} operation to the scaled image, and extract scene features by the third branch, using the whole ResNet50 encoder. Depth-wise convolution are then applied to reduce channel number. We denote the third part of feature $\mathbf{F_b}$.

By extracting illumination features $\mathbf{F_i}$ from holistic statistics, noise features $\mathbf{F_n}$ from low level representations, and blurry scene features $\mathbf{F_b}$ from high level extractors, we expect the features form complimentary representations, being effective in accurately predicting IRP values.

\begin{figure}[t]
\centering
\includegraphics[angle=0,width=0.5\textwidth]{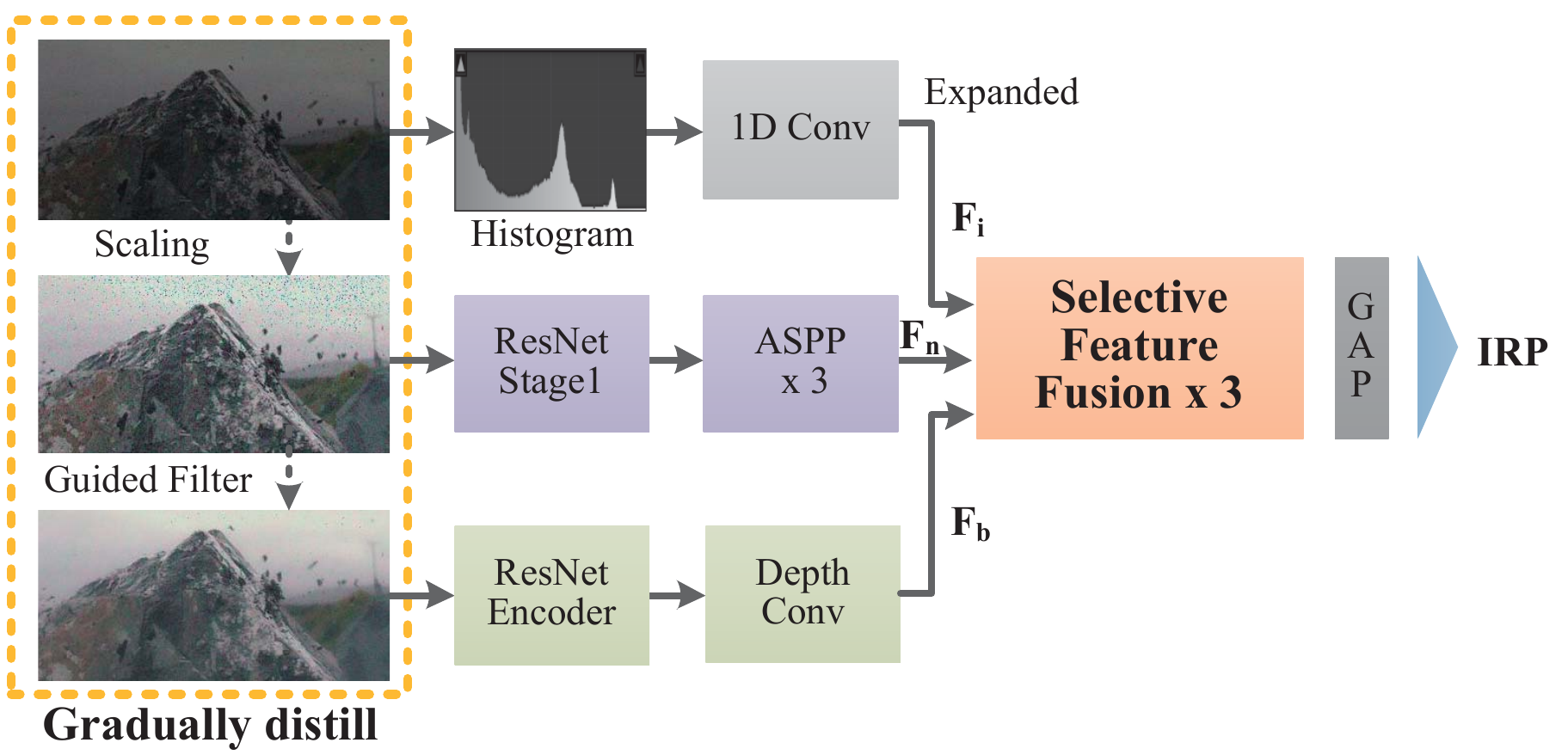}
\caption{The proposed model architecture for IRP prediction.}
\vspace{-0.6cm}
\end{figure}

\subsection{Selective Fusing Degradation Features}
\label{subsec:4.2}

As analysed above, when camera exposure setting varies, different kinds of distortions become determinant for IRP among the composite degradation. In order to dynamically adjust to dominant distortion factors, we propose to selective fuse the three parts of distortion features through a self-attention mechanism (also see supplementary for figure illustration), motivated by \cite{zamir2020learning}. 

Given three parts of extracted feature $\mathbf{F_i}$, $\mathbf{F_n}$ and $\mathbf{F_b}$, we first combine them through element-wise summation and global average pooling, to form a channel-wise feature representation $\mathbf{s}$: 
\vspace{-0.2cm}
\begin{equation}
    \mathbf{s} =  \rm{GAP}(\mathbf{F_i} + \mathbf{F_n} + \mathbf{F_b})
\end{equation}

\noindent
where $\rm{GAP}$ denotes global average pooling, $\mathbf{F_i}, \mathbf{F_i}, \mathbf{F_i} \in \mathbb{R}^{H\times W\times C}$and $\mathbf{s} \in \mathbb{R}^{1\times 1\times C}$.

We then apply a depth-wise convolution with squeeze ratio $r$ to $\mathbf{s}$, resulting a compact representation $\mathbf{z} \in \mathbb{R}^{1\times 1\times \frac{C}{r}}$. $\mathbf{z}$ are then fed into three parallel $1\times1$ convolution layers with expand ratio $r$ to get three feature indicators $\mathbf{u_1}$, $\mathbf{u_2}$ and $\mathbf{u_3}$, which are further re-weighted to attention activations by channel-wise softmax operation:
\vspace{-0.2cm}
\begin{equation}
    \mathbf{v_i}=\frac{e^{\mathbf{u_i}}}{\sum_{j}{e^{\mathbf{u_j}}}}
\end{equation}

Finally, the degradation features $\mathbf{F_i}$, $\mathbf{F_n}$ and $\mathbf{F_b}$ are adaptively selected by multiplying $\mathbf{u_1}$, $\mathbf{u_2}$ and $\mathbf{u_3}$ respectively. The selective fusion operation incorporates three kinds of degradation features and refines each of them to adjust to the variation of scene exposures, thus adapting well in the IRP prediction task.

\subsection{Regression to IRP Scores}

After distilling degradation components and extracting corresponding features, we repeatedly selective fuse the features by 3 times. The output features are then summed over and globally average pooled into a vector representation. Finally, three full connection layers are applied to regress the features into the IRP score. During training, we minimize $L1$ loss for optimization.

\section{Experiments}

In this section, to demonstrate the superiority of the proposed model, we evaluate IRP prediction accuracy on both synthetic and real world data. Due to the lack of models in the literature used for predicting the newly proposed IRP, we select existing IQA models as competitors.

\begin{table}[t]
\centering
\caption{IRP prediction accuracy comparisons on DS-IRP dataset.}
\smallskip
\footnotesize
\begin{tabular}{c|ll|ll}
\hline
\multirow{2}{*}{Model}          & \multicolumn{2}{c|}{Scene Average}                   & \multicolumn{2}{c}{Overall}                         \\ \cline{2-5} 
                                & \multicolumn{1}{c}{SRCC} & \multicolumn{1}{c|}{PLCC} & \multicolumn{1}{c}{SRCC} & \multicolumn{1}{c}{PLCC} \\ \hline
BRISQUE \cite{Mittal2012BRISQUE}                         & 0.3319                   & 0.3560                    & 0.1053                   & 0.1877                   \\
IL-NIQE \cite{Bovik2015ILNIQE}                        & 0.2631                   & 0.3330                    & 0.1869                   & 0.2051                   \\
HOSA \cite{Xu2016Blind}                           & 0.3360                   & 0.3448                    & 0.2269                   & 0.2014                   \\
DBCNN \cite{zhang2018blind}                          & 0.8022                   & 0.8008                    & 0.6903                   & 0.6956                   \\
KonCept512 \cite{hosu2020koniq}                 & 0.8892                   & 0.8984                    & 0.7536                   & 0.7839                   \\
HyperIQA \cite{su2020blindly}                        & 0.8483                   & 0.8578                    & 0.7383                   & 0.7550                   \\
Proposed                        & \textbf{0.9340}          & \textbf{0.9412}           & \textbf{0.8461}          & \textbf{0.8687}          \\ \hline
\end{tabular}
\label{tab:ds-irp}
\end{table}

\subsection{Evaluation on the DS-IRP Dataset}
\label{sec:5.1}

 We split the proposed DS-IRP dataset into a training subset, a validation subset and a testing subset, each containing 70\%, 10\% and 20\% images according to scene contents. All the competing models are trained following their default settings and the best performing models in validation set are selected for testing. During the evaluation, we calculate Spearman’s rank order correlation coefficient (SRCC) and Pearson’s linear correlation coefficient (PLCC) for comparison. We compute SRCC and PLCC within each individual scene in DS-IRP and report their average value, denoted as scene average. We also compute the two criteria on the whole test set to evaluate the overall model performance. As shown in Table \ref{tab:ds-irp}, the proposed model outperforms competitors by a large margin, indicating the effectiveness of the proposed architecture.

\begin{figure}
    \centering
    \subfloat[SRCC]{
        \begin{minipage}[t]{0.49\linewidth}
            \includegraphics[width=1\textwidth,trim= 12 0 12 0,clip ]{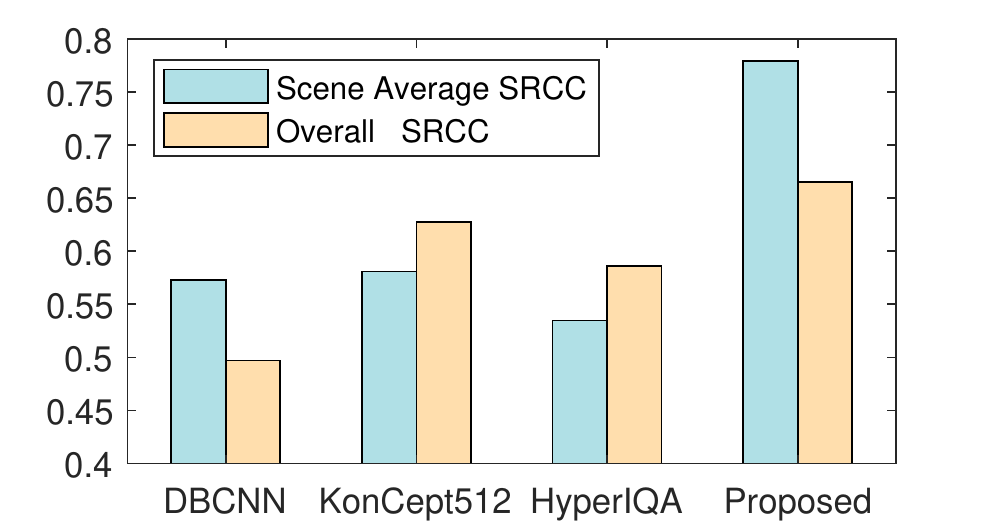}
        \end{minipage}
    }
    \hspace{-0.4cm}
    \subfloat[PLCC]{
        \begin{minipage}[t]{0.49\linewidth}
            \includegraphics[width=1\textwidth,trim= 12 0 12 0,clip]{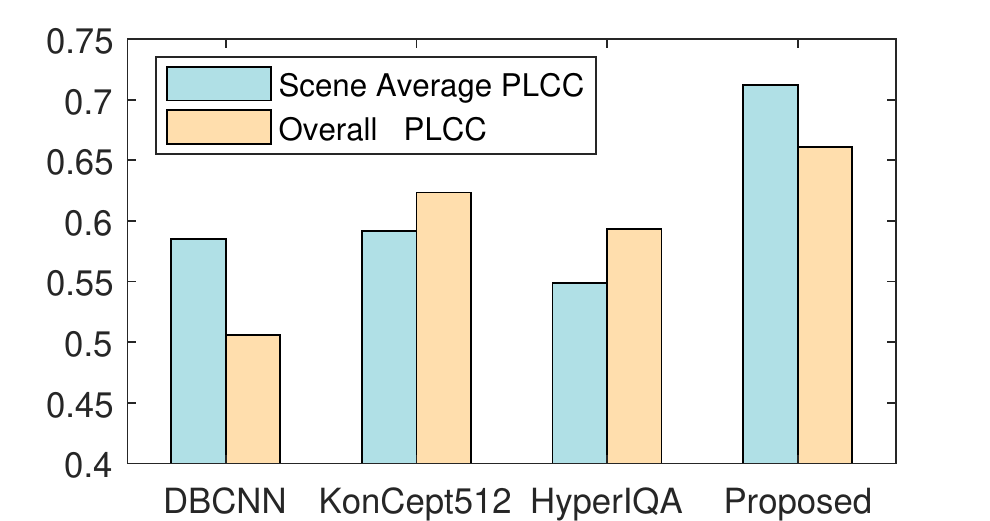}
        \end{minipage}
    }
    \vspace{-0.3cm}
    \caption{IRP prediction accuracy comparisons on real world data.}
    \vspace{-0.6cm}
    \label{fig:5.2}
\end{figure}

\subsection{Evaluation on Real World Data}
\label{sec:5.2}

We further compare model performances on images captured in real world scenarios. We first captured 20 scenes containing dynamic motions in real world, using 8 different exposures under each scene. The collection leads to 160 i-eps-converted-to.pdfmaging results in total, and we use models that are trained on the DS-IRP dataset to predict their IRP values.
Due to the lack of ground truth IRP to real world data, several pre-trained restoration methods \cite{chen2021hinet,zamir2021multi,zhang2019deep} are applied to process real world data and we further collected subjective perceptual scores from 40 participants to the restored results as substitutions of IRP values. We then calculate SRCC and PLCC among the model prediction scores and subjective scores, and show the results in Figure \ref{fig:5.2}. Similarly, the proposed model also outperformed competitors on the challenging real world data.

\begin{table}[ht]
\centering
\vspace{-0.2cm}
\caption{Ablation study for the proposed model.}
\vspace{-0.2cm}
\smallskip
\footnotesize
\begin{tabular}{c|ll|ll}
\hline
\multirow{2}{*}{Model} & \multicolumn{2}{c|}{Scene Average}                   & \multicolumn{2}{c}{Overall}                         \\ \cline{2-5} 
                       & \multicolumn{1}{c}{SRCC} & \multicolumn{1}{c|}{PLCC} & \multicolumn{1}{c}{SRCC} & \multicolumn{1}{c}{PLCC} \\ \hline
Baseline               & 0.8594                   & 0.8608                    & 0.7435                   & 0.7758                   \\
w/o illumination       & 0.9188                   & 0.9222                    & 0.8339                   & 0.8543                   \\
w/o blur               & 0.8886                   & 0.8992                    & 0.7814                   & 0.8188                   \\
w/o noise              & 0.8957                   & 0.9029                    & 0.7920                   & 0.8242                   \\
w/o selective          & 0.9335                   & 0.9385                    & 0.8278                   & 0.8476                   \\
Full                   & \textbf{0.9340}          & \textbf{0.9412}           & \textbf{0.8461}          & \textbf{0.8687}          \\ \hline
\end{tabular}
\vspace{-0.2cm}
\label{tab:ablation}
\end{table}

\subsection{Ablation Study}

We conduct ablation studies to validate the effectiveness of each model component. We remove the illumination, noise and blur branch separately in our model, to observe the effect of individual degradation features. We then remove the selective feature fusion in the model to validate the effectiveness of feature selection among composite distortions. We also evaluate model performance using a ResNet50 baseline, and the results are shown in Table \ref{tab:ablation}.

\section{Applications}

With the proposed IRP prediction model, in this section, we are able to demonstrate multiple applications that benefit from IRP prediction.

\begin{figure*}[htbp]
\centering
\subfloat{
\begin{minipage}[t]{0.75\linewidth}
   \includegraphics[width=1\textwidth,trim = 2 2 2 2,clip]{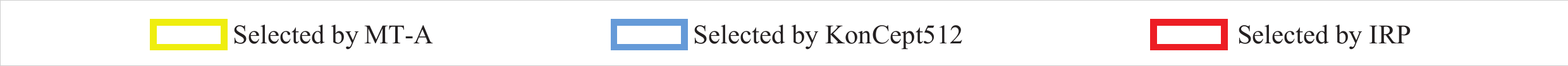}
\end{minipage}
}

\subfloat[28.49 dB]{
\begin{minipage}[t]{0.195\linewidth}
   \includegraphics[width=1\textwidth]{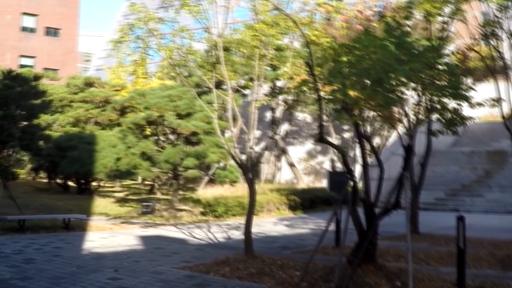}
   \includegraphics[width=1\textwidth]{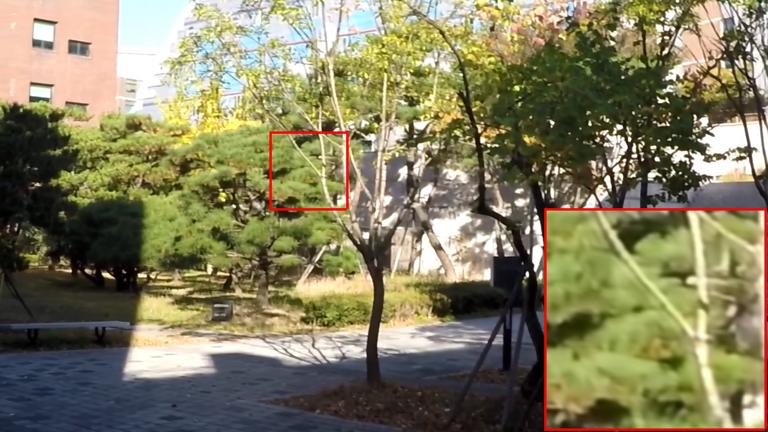}
\end{minipage}
}
\hspace{-2mm}
\subfloat[25.18 dB]{
\begin{minipage}[t]{0.195\linewidth}
   \includegraphics[width=1\textwidth]{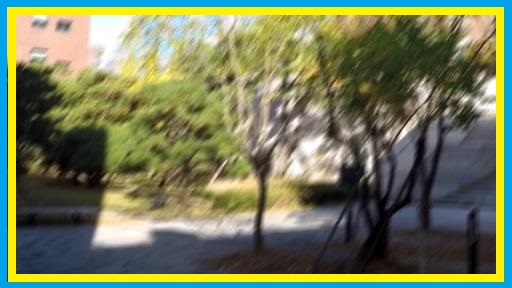}
   \includegraphics[width=1\textwidth]{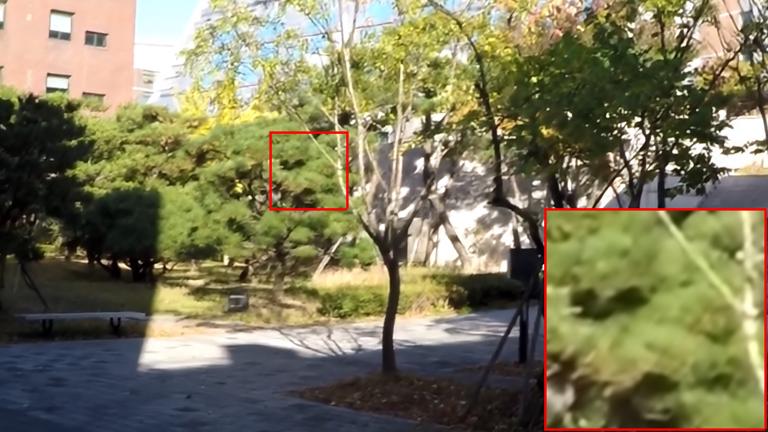}
\end{minipage}
}
\hspace{-2mm}
\subfloat[27.67 dB]{
\begin{minipage}[t]{0.195\linewidth}
   \includegraphics[width=1\textwidth]{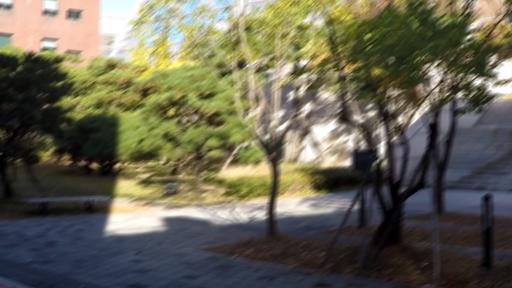}
   \includegraphics[width=1\textwidth]{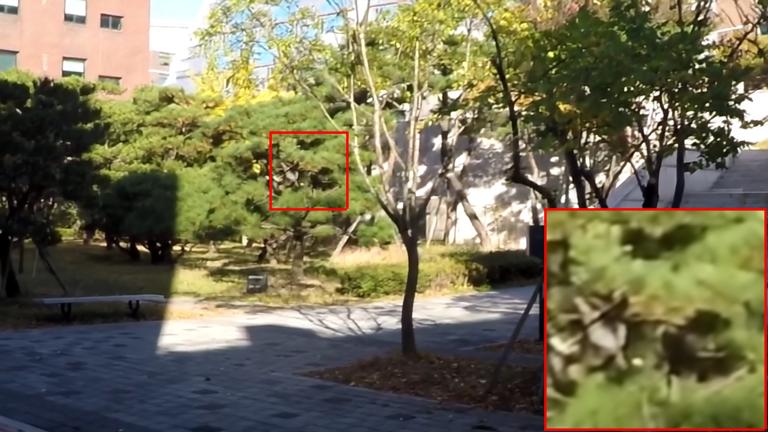}
\end{minipage}
}
\hspace{-2mm}
\subfloat[27.64 dB]{
\begin{minipage}[t]{0.195\linewidth}
   \includegraphics[width=1\textwidth]{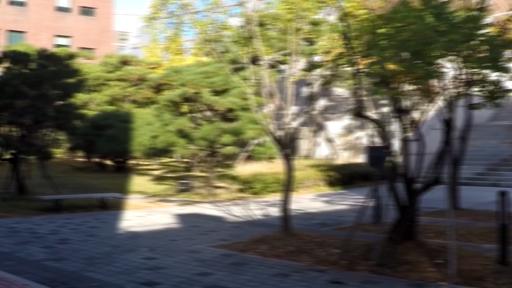}
   \includegraphics[width=1\textwidth]{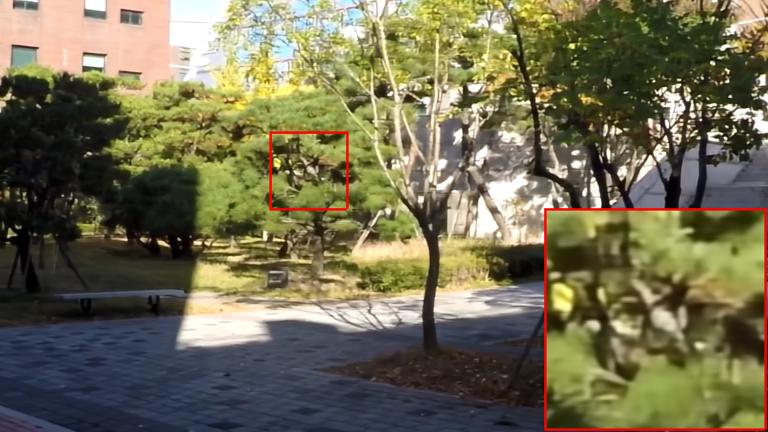}
\end{minipage}
}
\hspace{-2mm}
\subfloat[\textbf{31.09 dB}]{
\begin{minipage}[t]{0.195\linewidth}
   \includegraphics[width=1\textwidth]{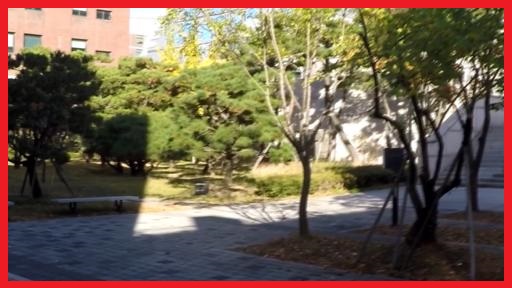}
   \includegraphics[width=1\textwidth]{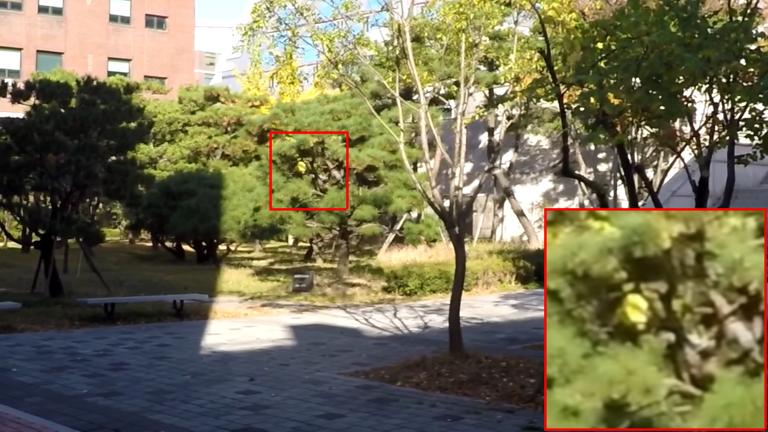}
\end{minipage}
}

\subfloat[\textbf{32.38 dB}]{
\begin{minipage}[t]{0.195\linewidth}
   \includegraphics[width=1\textwidth]{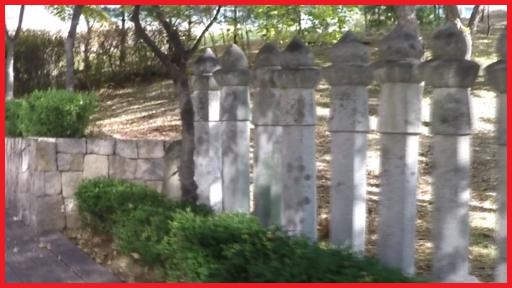}
   \includegraphics[width=1\textwidth]{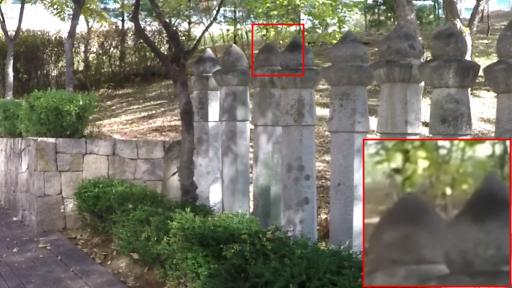}
\end{minipage}
}
\hspace{-2mm}
\subfloat[31.22 dB]{
\begin{minipage}[t]{0.195\linewidth}
   \includegraphics[width=1\textwidth]{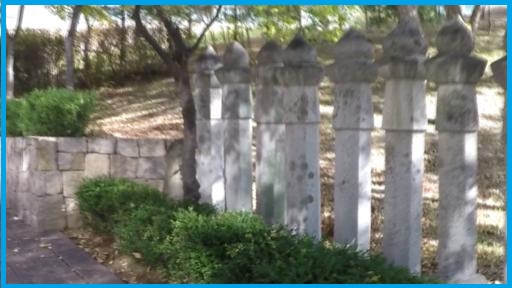}
   \includegraphics[width=1\textwidth]{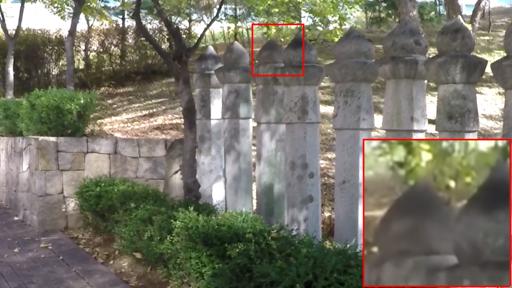}
\end{minipage}
}
\hspace{-2mm}
\subfloat[30.75 dB]{
\begin{minipage}[t]{0.195\linewidth}
   \includegraphics[width=1\textwidth]{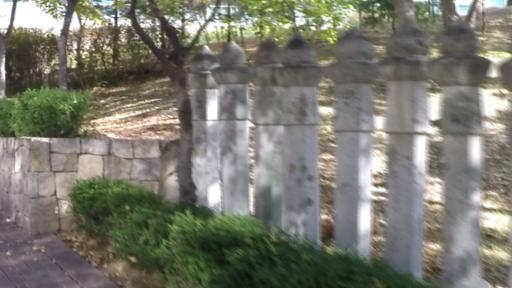}
   \includegraphics[width=1\textwidth]{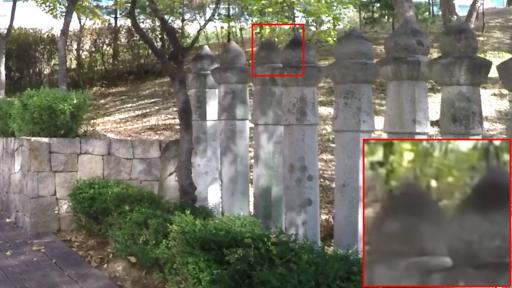}
\end{minipage}
}
\hspace{-2mm}
\subfloat[29.51 dB]{
\begin{minipage}[t]{0.195\linewidth}
   \includegraphics[width=1\textwidth]{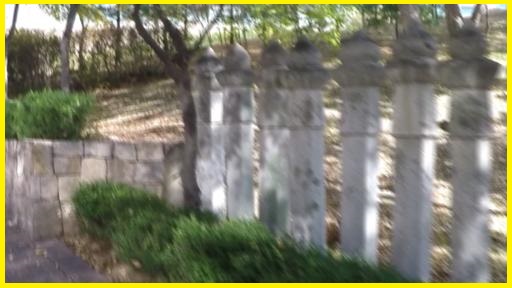}
   \includegraphics[width=1\textwidth]{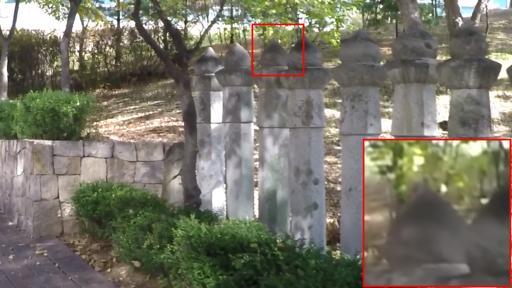}
\end{minipage}
}
\hspace{-2mm}
\subfloat[27.47 dB]{
\begin{minipage}[t]{0.195\linewidth}
   \includegraphics[width=1\textwidth]{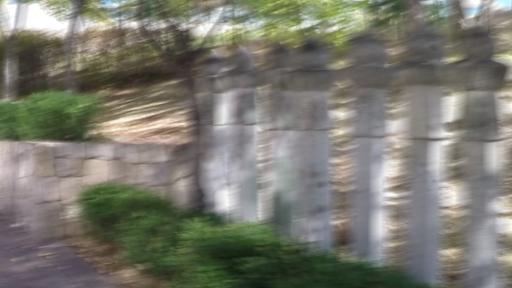}
   \includegraphics[width=1\textwidth]{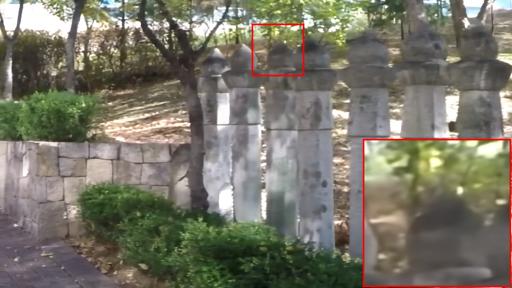}
\end{minipage}
}
\caption{We show how IRP serves as a filtering principle on the GoPro dataset. Compared with KonCept512 and MT-A, IRP prediction correctly select the most valuable frame in a sequence for efficiently restoring.}
\vspace{-0.6cm}
\label{fig:6.1}
\end{figure*}

\subsection{Filtering Principle for Efficient Processing}

There exist circumstances such as autonomous driving and robot vision navigation, where images captured in dynamic scenes require to be processed and restored in real-time. 
To meet time and computation requirement, image filtering strategy becomes a solution for efficient processing. Specifically, among the continuous frames showing similar scene contents, only valuable frames are selected and processed to avoid overmuch raw data. In this way, both time cost and overall restored image quality could be improved.

Under the circumstance, IRP prediction can serve as a filtering principle, and we evaluate its effectiveness on the real world deblurring dataset GoPro \cite{nah2017deep}. Specifically, we split the GoPro test set into 105 groups containing 10 continuous frames inside each group. We then select the best frame inside each group according to the predicted IRP values, and process the selected frame by DMPHN\cite{zhang2019deep}. We evaluate overall model complexity, average time consumption, restoration quality and the best frame selection accuracy in Table \ref{tab:gopro}. To make comparisons, we also adopt two IQA models, which are trained on their own proposed IQA datasets as filtering principles, including KonCept512 trained on the authentically distorted IQA dataset KonIQ-10k \cite{hosu2020koniq} , and MT-A trained on the smartphone photography and quality dataset SPAQ \cite{fang2020perceptual}. We also show qualitative comparisons in Figure \ref{fig:6.1}.

\begin{table}[th]
\centering\
\vspace{-0.4cm}
\caption{Model evaluations on the GoPro dataset as a frame filtering principle.}
\smallskip
\footnotesize
\begin{tabular}{c|ccccc}
\hline
Model      & FLOPs(G)       & Time(s)         & PSNR            & SSIM            & Accuracy        \\ \hline
DMPHN      & 1099.35        & 0.0521          & 30.453          & 0.9022          & -               \\
KonCept512 & 152.234        & 0.0342          & 31.849          & 0.9263          & 49/105          \\
MT-A  & \textbf{9.394} & \textbf{0.0205} & 31.155          & 0.9116          & 34/105          \\
IRP        & 41.895         & 0.0244          & \textbf{32.140} & \textbf{0.9351} & \textbf{57/105} \\ \hline
\end{tabular}
\vspace{-0.2cm}
\label{tab:gopro}
\end{table}

From Table \ref{tab:gopro}, we found that for all three filtering models, the average restored image quality improved. This demonstrates both the feasibility of the image filtering strategy and the potential extended usages of existing IQA models. Meanwhile, among the competing models, both restored image quality and frame selection accuracy perform best through our IRP prediction model. Although our model is trained on the synthetic DS-IRP dataset, it outperforms competitors which collect real world images and subjective scores for model training. The result further proves the IRP prediction as a superior filtering principle for image restoration applications.

\begin{table}[th]
\centering
\caption{Performance comparison on the proposed IRP as an auxiliary guidance for training restoration models.}
\smallskip
\footnotesize
\begin{tabular}{ccc}
\hline
Model        & PSNR            & SSIM            \\ \hline
CBD baseline & 40.227          & 0.9793          \\
CBD full     & 40.722          & 0.9818          \\
CBD + IRP    & \textbf{40.771} & \textbf{0.9821} \\ \hline
\end{tabular}
\label{tab:sidd}
\end{table}

\begin{figure}[th]
\centering
\subfloat[Noisy]{
\begin{minipage}[t]{0.23\linewidth}
   \includegraphics[width=1\textwidth]{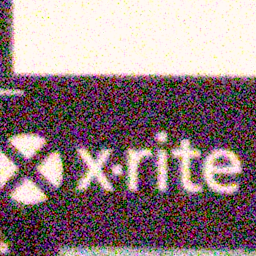}
   \includegraphics[width=1\textwidth]{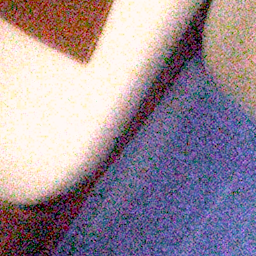}
\end{minipage}
}
\hspace{-1.5mm}
\subfloat[CBD base]{
\begin{minipage}[t]{0.23\linewidth}
   \includegraphics[width=1\textwidth]{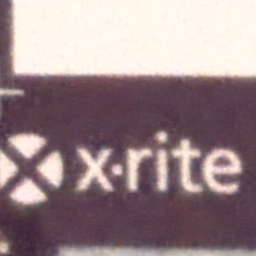}
   \includegraphics[width=1\textwidth]{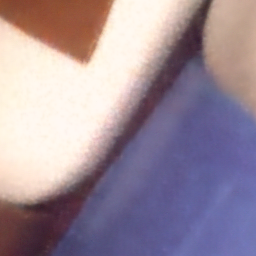}
\end{minipage}
}
\hspace{-1.5mm}
\subfloat[CBDbase+IRP]{
\begin{minipage}[t]{0.23\linewidth}
   \includegraphics[width=1\textwidth]{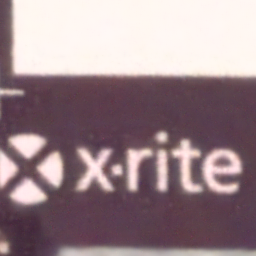}
   \includegraphics[width=1\textwidth]{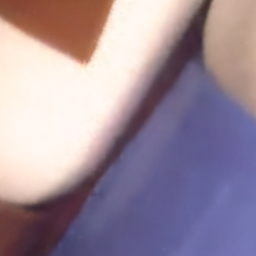}
\end{minipage}
}
\hspace{-1.5mm}
\subfloat[Ground Truth]{
\begin{minipage}[t]{0.23\linewidth}
   \includegraphics[width=1\textwidth]{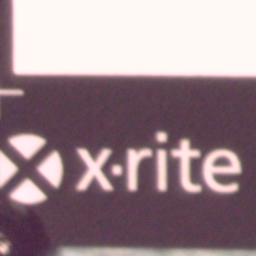}
   \includegraphics[width=1\textwidth]{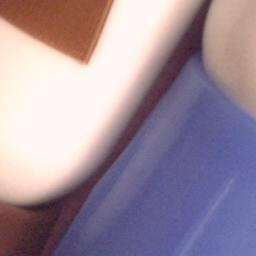}
\end{minipage}
}
\caption{Qualitative comparisons of IRP as an auxiliary guidance for restoration models.}
\vspace{-0.8cm}
\label{fig:6.2.2}
\end{figure}

\begin{figure*}[t]
\centering
\subfloat[Auto-exposure]{
\begin{minipage}[t]{0.24\linewidth}
   \includegraphics[width=1\textwidth]{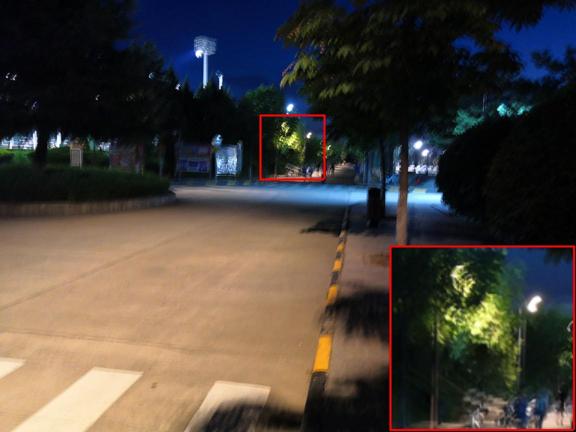}
   \includegraphics[width=1\textwidth]{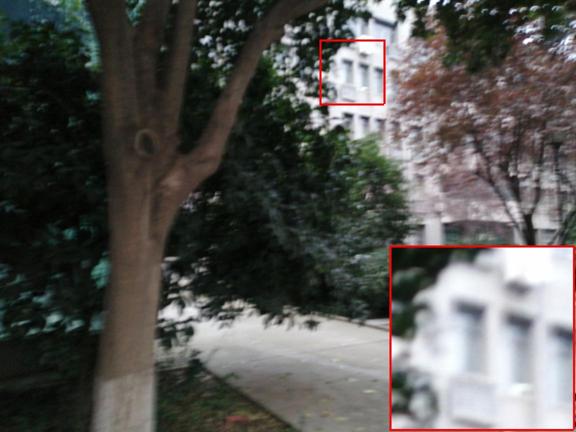}
   \includegraphics[width=1\textwidth]{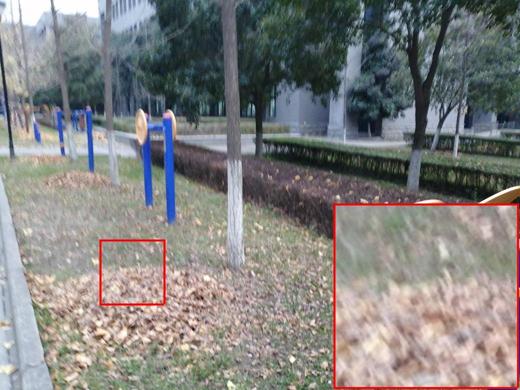}
\end{minipage}
}
\subfloat[IRP-exposure]{
\begin{minipage}[t]{0.24\linewidth}
   \includegraphics[width=1\textwidth]{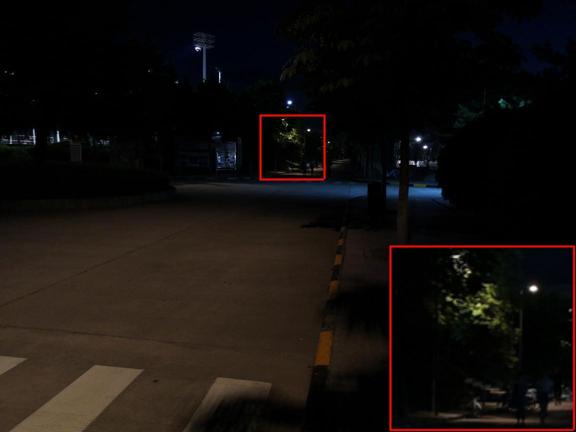}
   \includegraphics[width=1\textwidth]{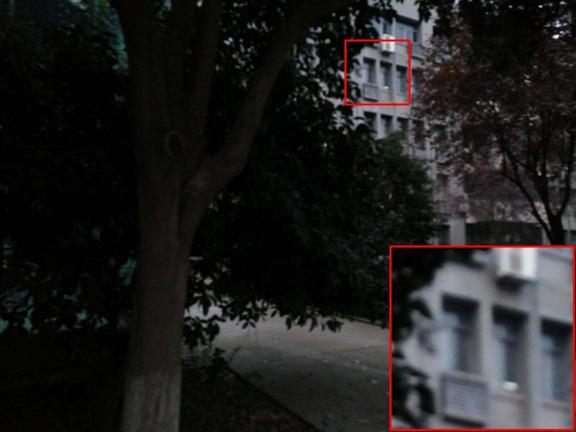}
   \includegraphics[width=1\textwidth]{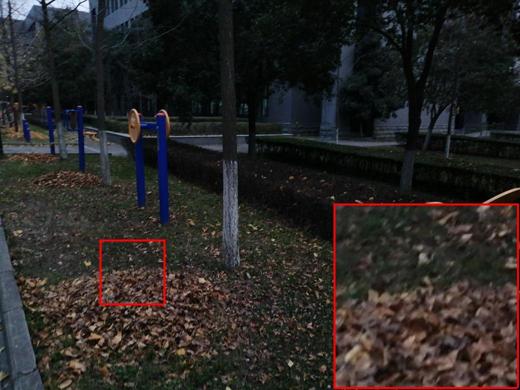}
\end{minipage}
}
\subfloat[Auto-exposure restored]{
\begin{minipage}[t]{0.24\linewidth}
   \includegraphics[width=1\textwidth]{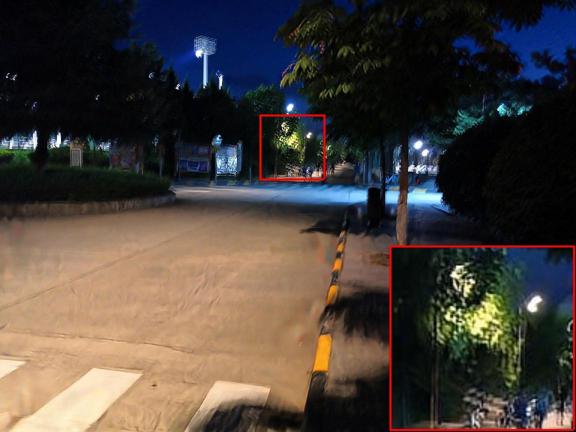}
   \includegraphics[width=1\textwidth]{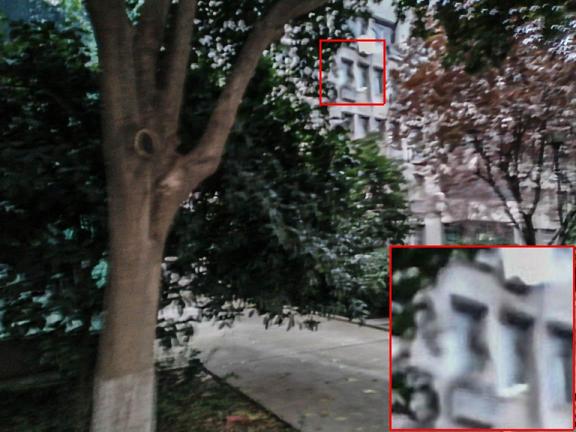}
   \includegraphics[width=1\textwidth]{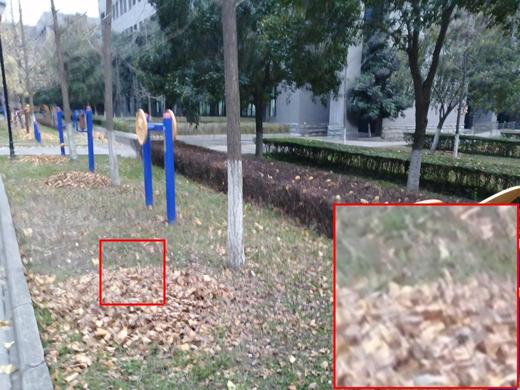}
\end{minipage}
}
\subfloat[IRP-exposure restored]{
\begin{minipage}[t]{0.24\linewidth}
   \includegraphics[width=1\textwidth]{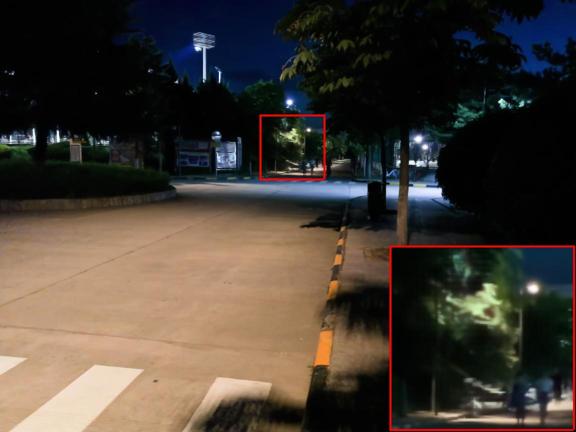}
   \includegraphics[width=1\textwidth]{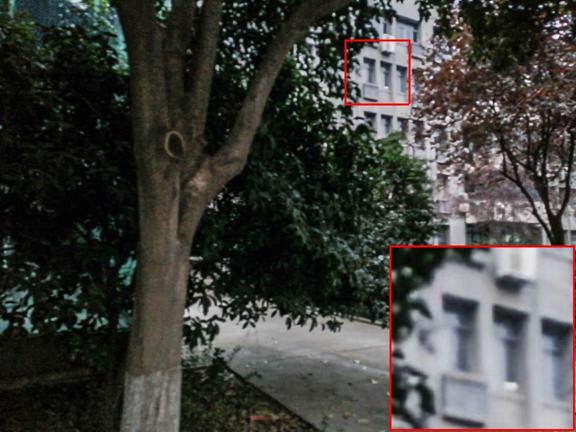}
   \includegraphics[width=1\textwidth]{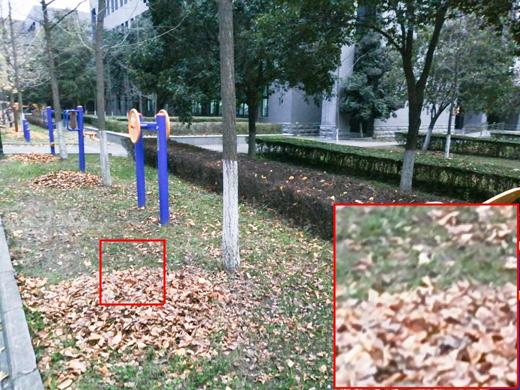}
\end{minipage}
}
\caption{Comparison of IRP for optimizing exposure settings against conventional auto-exposure settings of camera.}
\vspace{-0.6cm}
\label{fig:6.3}
\end{figure*}

\subsection{Auxiliary Guidance for Restoration Models}

We also explore if IRP can also provide guidance for training adaptive restoration models. The underlying assumption is that by feeding both the image and its IRP information to restoration models, they are able to distinguish images as easy or hard samples, thus learning adaptive mappings for images showing various restoration potentials.

To validate the assumption, we select small RGB data from the image denoising dataset SIDD \cite{abdelhamed2018high} for evaluation. Given an input image, we extract its IRP features through the proposed model trained in Section \ref{sec:5.1} and use a $1\times1$ convolution layer to adjust the feature map into 3 channels. The feature map, concatenated with the original image, is then fed into the restoration model for either training or testing. We compare model performance with CBDnet\cite{guo2019toward}, which also uses an auxiliary subnetwork, especially trained on the mixture of synthetic and real noisy images, to estimate image noise levels for restoration guidance. We also compare model performance when trained without any auxiliary guidance, denoted as CBDnet baseline, and show the results in Table \ref{tab:sidd}, Figure \ref{fig:6.2.2}. It can be found that by adding an auxiliary guidance, both CBDnet and IRP boost the image restoration baseline. Moreover, by extracting IRP features, restoration models are even more benefited than CBDnet. Though IRP is proposed to deal with composite distortions and we do not explicitly train it on real noisy images, we found it achieved impressive performance on the denoising task. Furthermore, since IRP features are extracted separately from restoration models, it is expected that IRP prediction can serve as a plug-and-play module in improving many other image restoration tasks.

\subsection{Indicator for Optimizing Camera Settings}

In the above subsections, we show IRP applications on single deblurring or denoising tasks, in this subsection, we further show its usages under real world composite distortions. We show the potential usage that applying IRP in optimizing camera exposure settings in real world dynamic imaging scenarios. As conventional auto-exposure settings in existing camera devices guarantee sufficient illumination as a priority, in dynamic scenes, whether the captured image suits its best for restoration cannot be promised. We illustrate the application in Figure \ref{fig:6.3}, where we show real world images collected from Section \ref{sec:5.2}, selected by camera auto exposures and by IRP predictions. Both originally distorted images and the restored results are presented for qualitative comparisons. As can be seen, the conventional auto-exposure setting tends to capture images with sufficient illumination, they do not lead to satisfying restored quality. As a comparison, by predicting the origin images' IRP values, the selection leads to restoration results showing more satisfying visual quality.

\section{Conclusion}

In this paper, we propose IRP, a novel image attribute measuring its potential power that can be explored for restoration. We first established a DS-IRP dataset and explored the properties of IRP. Based on the analysis, we further proposed a deep model which gradually distills and selective fuse degradation features to accurately predict IRP. Experimental evaluations demonstrate the effectiveness of the proposed model architecture. Finally, with the IRP prediction model, we are able to apply it in various image restoration related tasks. The IRP has shown its potential usages in filtering valuable frames for efficient processing, providing extended guidance for restoration models, and even optimizing camera settings when capturing images under dynamic scenarios.

%-------------------------------------------------------------------------

%%%%%%%%% REFERENCES
{\small
\bibliographystyle{ieee_fullname}
\bibliography{egbib}
}

\end{document}